\documentclass[11pt]{article}

\usepackage[utf8]{inputenc}
\usepackage[T1]{fontenc}
\usepackage[english]{babel}
\usepackage{amsmath,amssymb,amsfonts}
\usepackage{graphicx}
\usepackage{geometry}
\usepackage{booktabs}
\usepackage{array}
\usepackage{hyperref}
\hypersetup{
    colorlinks=true,
    linkcolor=blue,
    citecolor=blue,
    urlcolor=blue
}
\usepackage{caption}
\usepackage{float}

\title{\textbf{A Layer Importance Metric for Quantization Accounting for the Speed--Quality Trade-off in Autoregressive Models}}

\author{A.K. Safronov \\ \small safro@sfedu.ru}
\date{}

\begin{document}

\maketitle

\begin{abstract}
Small large language models (sLLMs) are now deployed on devices with limited memory and compute budget. Within the autoregressive setting, inference is memory-bandwidth bound: typical homogeneous quantization is typically detrimental to such models due to their compactness, as their architectures do not retain much redundancy, and in particular a handful of components display extremely poor tolerance to lower precision.

This paper introduces a composite metric that combines two orthogonal assessment criteria for such components: the information retention after quantization, captured by a normalized coefficient, and the gains in throughput that can be expected from accelerating individual layers, analytically estimated using a roofline-based latency model. Using an architectural profiling benchmark on Gemma 3 1B, we identify the Feed-Forward Network (FFN) blocks and the embedding matrix as the most promising candidates for accelerated inference, and we estimate for each candidate a normalized quality score, computed via simulated quantization, and a normalized speed score, computed without any model execution. The metric uses a single prioritization coefficient to combine the two scores into a composite value, which can be tuned to prioritize speed vs quality according to a specific application’s needs. The metric is general and can be applied to blocks of layers (such as FFNs in the transformer), their internal projections (such as the feed forward network within a transformer block), or single transformer layers, with no modifications to the computation.

We evaluated our method on multiple model architectures, measuring around 4\% error in predicted speedups on average, and demonstrated that it enables significantly better resource allocation decisions than existing approaches, which usually rely on extensive evolutionary search strategies, domain-specific inference accelerators, or require access to expensive Shapley-value approximations. In short, our analytical approach makes sLLM quantization both faster, easier, and more intuitive, turning a trial-and-error process into a mathematical exercise.
\end{abstract}


\section{Introduction}

The current practice in large language model (LLM) design appears to be contradictory, in that while model architectures become increasingly complex, their physical realization is subject to rather strict limitations imposed by memory bus widths, battery capacity, or the available silicon area. AI assistants are now omnipresent in smartphones and other local machines where they perform a variety of non-critical tasks such as content creation and web browsing~\cite{ref23}. Yet even in this setting, when an LLM is used for generic-purpose assistance, the emphasis is placed on the model's ability to operate in a memory-constrained environment rather than perform hundreds of billions of operations per second; put otherwise, a compactly represented LLM is significantly more valuable than a large one at the same level of functional complexity. Now, the critical question is: how to achieve a high degree of model reduction while preserving the response quality, as witnessed by the recent advances in knowledge distillation and parameter sharing.

It is not accidental that currently available small LLMs (Llama 3 1B/3B, Gemma 3 1B, Qwen 3 1.7B, etc.) are the result of aggressive model distillation: a large-scale LLM possesses ample resources to withstand the pruning of its parameter count, which has little effect on its accuracy, while an sLLM (small LLM) does not have these advantages. If post-training quantization (PTQ) is applied to an sLLM, it may not produce acceptable results: the internal structure of the model remains intact, but its behavioral patterns are altered, so that the response quality decreases substantially.

Another issue is that modern accelerators are often memory-bandwidth-constrained~\cite{ref3}: the majority of the time spent in autoregressive decoding is occupied by offloading weights from video memory, not performing arithmetic operations, which also appears to be the case when naively implementing the attention mechanism~\cite{ref16}.

Thus, the utilization of a GPU in this scenario is significantly lower than could be the case, and the user must wait for the next token to be generated.

A uniform reduction in the precision of weights and activations, applicable to each layer of the model, is not the only possibility; indeed, different network layers may respond differently to such a reduction. Some layers are relatively inert and can safely undergo aggressive quantization without noticeable effects on the quality of generation, while others store crucial information that should be left untouched. Several methods have been developed to date that allow one to quantify different elements of the model with varying degrees of precision. However, each of them is limited by certain design decisions: CARVQ~\cite{ref4} and IMPQ~\cite{ref6} utilize multi-level vector quantization or require extensive computation to estimate each component's contribution to the total model performance using Shapley-value estimation~\cite{ref21}, while HAPM~\cite{ref5} requires specialized inference software to demonstrate its effectiveness. Thus, none of these approaches provides a practical solution for easily identifying the set of layers within an sLLM that could benefit from higher precision.

The paper suggests that in order to achieve the best possible trade-off between performance and quality, one must move away from uniform quantization and instead attempt to find groups of operations within the model that contribute the most to its quality. In other words, it is more useful to spend computational resources (measured in FLOPS) on operations that preserve information rather than discard it.

We propose a method by which one can assess the quality of an sLLM quantization configuration using a single metric that estimates the cost-quality trade-off. The paper reports the following results:
\begin{itemize}
    \item profiling the Gemma 3 1B architecture to identify the blocks that contribute most to inference latency, and justifying the choice of the FFN and embedding matrix as priority quantization targets;
    \item obtaining a normalized quality score for the FFN and Embedding blocks via the SQNR coefficient using the SA-PTQ operator;
    \item obtaining a normalized speed score for each block based on an analytical roofline model, without re-running the model;
    \item combining both scores into an integrated metric with a priority parameter $\alpha$, allowing the balance between speed and quality to be shifted depending on the requirements of a specific task.
\end{itemize}

The analysis presented confirms the applicability of the metric on standard hardware without quantizing the model, and substantiates the possibility of extending it to layer-wise heterogeneous quantization.

\textbf{Goal of the present work.} The goal is to develop an analytical metric for estimating layer importance that accounts for the trade-off between inference speed and generation quality in autoregressive models. This tool is intended to identify structural blocks and layers whose quantization will provide the maximum hardware gain at minimal information loss.

\subsection{Domain analysis}

To justify the necessity of the methodology for the choice of blocks presented in the paper, it is essential to explain the technological context of optimization. Namely, the common industry practice which allows to reduce VRAM usage and thus accelerate inference without retraining is model quantization - a process of reducing the bit width of weights to 4 or 8 bits. It goes without saying that quantization is never the sole optimization step - it rather represents the final stage when most of the optimizations are already performed. And the issue with optimization is not quantitative - the tooling for quantitative analysis is lacking. In other words, a significant amount of blocks can be subjected to quantization without affecting the performance of the model, while some blocks cannot. This is why quantization as a concept is not the ultimate goal but only the stage that requires analysis. And this is the analysis that the given paper provides with, which is why quantization is framed as a task to solve and not a goal to achieve.

It is vital to understand that ultimately, the objective of the proposed metric is to prioritize model blocks and layers in terms of information robustness. This way, all blocks that require high precision to maintain their functional capabilities are separated from those that allow for aggressive optimization with little loss in terms of accuracy. And this prioritization is ultimately the basis for heterogeneous models which can provide the most significant speed gains with the least losses in performance.

\subsection{Related work}

Small language models have enabled to run LLMs on the end-user device. However, due to the nature of the Transformer architecture, generating tokens is a slow process, and an unquantized model is often too slow to be useful~\cite{ref22}.

Quantization, which reduces the bit-width of weights, is the standard method to tackle the problem. However, it is not trivial to apply existing methods to smaller models.

Compared to large models, small ones are more information-dense: they do not have the luxury of discarding some weights as not useful. As a result, applying the same compression to every layer harms the usefulness of the model more than its size.

Several papers addressed this issue from different angles. CARVQ aims to tackle the problem of embedding matrix size by compressing it to around 1.6 bits using a corrective adaptor and group residual vector quantization~\cite{ref4}. The authors note that, after PTQ of the transformer itself, the embedding matrix comprises the largest portion of the model (52\% for LLaMA-3.2-1B)~\cite{ref4}, so it should be compressed aggressively. CARVQ requires lookup tables and nonlinear transformations to be applied at inference time, which complicates the integration with other code. The metric proposed in this paper, in contrast, utilizes conventional scalar quantization (SA-PTQ) and measures the SQNR of the compressed weights, without requiring any additional structures. As a result, it can be used in existing tools such as llama.cpp.

HAPM paper~\cite{ref5} suggests another approach, prunning the weights of the language models on mobile phones with hardware-aware latency modeling. They introduce the concept of real latency sensitivity, which is linked directly to the accuracy drop of a pruned model. The structure of the resulting model becomes sparse, requiring specialized inference machinery to attain the reported speed gains. The $S_n$ metric from the current paper does not depend on model structure and can be applied to any model. It measures the information contained in the weights, normalized by the time spent on processing them ($\Delta L$). It achieves the desired speed gains by eliminating lower-ranked layers outright rather than changing their structure, and can utilize existing accelerators.

IMPQ paper~\cite{ref6} analyzes interactions between layers using Shapley values from game theory to determine the importance of different weight tensors. While it provides theoretical justification for the layer ranking, the computation of the Shapley values is expensive to apply to models with tens of billions of weights, as in the LLaMA series. $S_n$, on the other hand, only needs to perform a relatively quick analytical probing of the model to approximate the importance of each layer. This allows to spend less time on model analysis and focus on finding the best compression option.

QRazor paper~\cite{ref7} combines two-step weight and activation compression with Significant Data Razoring to compress the LLM by removing the least-significant bits. They modify the Attention blocks, but the authors claim that their changes to the performance are minimal. Unlike QRazor, which modifies the Attention blocks directly, this work focuses primarily on FFN and Embedding: efficient attention variants such as GQA~\cite{ref20} already reduce its memory footprint architecturally to some extent, and our tests (Section 2) showed that Attention contributes comparatively little to per-token latency in the configurations we profiled, making it a lower priority target here. QRazor uses a heuristic based on amplitudes of the weights to rank the importance of bits, while $S_n$ uses an energy-based SQNR metric to approximate the value of each bit more accurately. This allows to select the most valuable bits to keep, providing the flexibility to choose the desired generation quality/bit-width balance.

\subsection{Problem statement}

The goal of the present study is to develop a metric for evaluating neural network components that allows ranking them by significance for inference acceleration while preserving generation quality. Formally, the task can be represented as a search for a heterogeneous quantized model configuration that provides maximum performance under a given quality-loss threshold.

\textbf{Formal description of parameters.} Let the original model $M$ be represented as an ordered set of components $\{C_1, C_2, \ldots, C_n\}$, where a component may be an entire structural block (FFN, Embedding), an individual projection within a block ($W_{gate}, W_{up}, W_{down}$), or a separate transformer layer. The quantization process consists in mapping each layer to the space of admissible bit widths $B = \{b_4, b_8, b_{16}\}$. The state of the system is characterized by the following functions:

\begin{itemize}
    \item $L(L_i, b)$ -- the latency function, defining the inference execution time of block $L_i$ at bit width $b$;
    \item $Q(M, \{b_1, \ldots, b_n\})$ -- the integral quality indicator (functional robustness), characterizing the deviation of the quantized model's output signal from the reference.
\end{itemize}

To find the optimal bit-width distribution $\{b_1, b_2, \ldots, b_n\}$, we formulate an optimization problem. The objective function, aimed at minimizing total inference time (the efficiency criterion), is given by (1).

\begin{equation}
\sum_{i=1}^{n}\left( L_{MLP}^{(i)}(b_i) + L_{Emb}(b) \right) \longrightarrow \min
\end{equation}

At the same time, the system is subject to a boundary condition (2), defining the maximum admissible degradation of the information signal (the quality constraint).

\begin{equation}
SQNR_{total} \geq SQNR_{baseline} - \epsilon
\end{equation}

where expression (1) defines the efficiency criterion, and inequality (2) fixes the maximum admissible signal-energy loss $\epsilon$, expressed in decibels (dB).

\begin{figure}[H]
    \centering
    \includegraphics[width=0.75\textwidth]{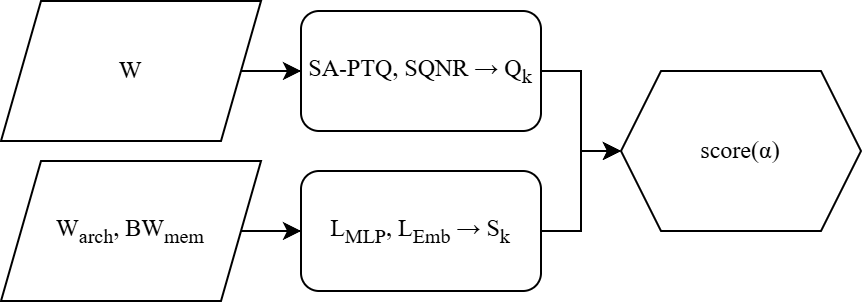}
    \caption{Algorithmic scheme for constructing the selectivity metric $score(\alpha)$. \emph{Source: compiled by the author.}}
    \label{fig:scheme}
\end{figure}

The result of this formalization is the development of an integral selectivity criterion $score(\alpha)$ (Fig.~\ref{fig:scheme}).

This criterion combines two normalized components -- the information preservation of a component $Q_k$ and its speed potential $S_k$ -- through the priority parameter $\alpha$, allowing a heterogeneous architecture to be formed that is balanced with respect to speed and generation quality.

\section{Formalizing the Latency Parameters of FFN and Embedding}

For an effective compression-evaluation metric, one has to move away from an abstract consideration of weight counts to the concrete impact each component’s size has on the resulting inference speed. And indeed, when speaking of sLLMs, the relevant metric is not the raw power, but the latency introduced by reading data from memory. A cost metric, discussed here, aims to provide a means of precise prediction by answering the question of what ``gain’’ in performance, measured in seconds saved, is projected from reducing the bit-width of a particular layer. This metric turns the vague question of how many layers to quantize into an exacting calculation, the parameters of which are defined by the desired acceleration and the quality of the signal preserved.

Beforehand, however, a profiling operation was performed on the Gemma 1B model to identify the major contributors to token-generation latency. The metric of choice, for this particular study, was simply the token generation speed, as it defines the applicability of a quantized model in practice. The blocks under review were 8-bit quantized to identify the primary sources of overhead, which were then isolated for further analysis.

The operations of profiling and identifying computational dominants were performed in Google Colab with T4 GPU using the resources of the llama.cpp toolkit. In particular, the \texttt{llama-quantize} utility, which allows for changing the bit-width of individual blocks by means of typecasting particular tensors, was used. Gemma 3 1B was quantized to Q8\_0 for the relevant parameter groups: FFN, Attention mechanism layers (Q, K, V, Out) and Embedding matrix, with F16-weights model serving as the reference point. The metrics for each configuration were gathered automatically with the help of the \texttt{llama-cpp-python} library with at least three runs for each set of parameters to eliminate hardware variation noise.

The characteristics of architectural performance were defined with the consideration of three factors: token generation speed (tokens/s), context processing latency and dynamic VRAM consumption. The latter, in particular, was evaluated by means of direct reading of the VRAM utilization percentages from the \texttt{nvidia-smi} utility with 0.05s sampling frequency throughout the duration of calculations. Diversified input sequences were used for the calculations to observe the behavior of the model under both light (short queries) and heavy (8192 tokens, typical for gaming sessions) loads.

The results of the profiling runs (see Table~\ref{tab:profiling}) confirm the nonlinear nature of the relationship between the bit-width of model sections and their impact on the inference speed. In particular, the analysis demonstrates that the contribution of the FFN blocks and the Embedding matrix to the overall performance is disproportionately high compared to the input dimensions: thus, the speed gain from reducing their bit-depth from F16 to Q8 is substantially larger for both than for Attention layers.

On the other hand, the comparison of the configurations with reduced precision of FFN and Embedding (``Q8 Q8 Q8’’) and the baseline (``Q8 ffn\_all=F16’’) show that the omission of these blocks from quantization results in significantly decreased performance of the 8-bit model. In particular, for the short context length, the token generation speed decreases to 72.3 tok/s for the latter configuration, compared to the fully quantized alternative. And while the contribution of the embedding layer is less obvious, the drop in throughput for completely unquantized Embedding is still substantial.

By comparison, the impact of the Attention layers is significantly lower: even with a similar reduction in precision (from F16 to Q8), the loss of performance from omitting these blocks is smaller for both context lengths. Altogether, the observations confirm the hypothesis that in the Gemma 3 1B architecture, the FFN blocks and the Embedding matrix are the critical performance factors. Their contribution to the overall latency dominates over other sources of overhead and, therefore, should be prioritized when selecting layers for quantization in a local deployment.

\begin{table}[H]
\centering
\caption{Performance measurements for different quantization variants of Gemma 3 1B blocks (FFN/Attn/Emb) at context lengths of 10, 50, and 100 tokens.}
\label{tab:profiling}
\small
\begin{tabular}{lrrrrr}
\toprule
Variant (FFN/Attn/Emb) & short & dnd\_10 & dnd\_50 & dnd\_100 & size, MB \\
\midrule
Q8 only attn\_v          & 62.3 & 63.7 & 61.3 & 58.0 & 1999 \\
Q8 only attn\_kv         & 63.1 & 63.9 & 58.5 & 59.1 & 1992 \\
Q8 only attn\_k          & 64.4 & 60.8 & 61.4 & 58.0 & 1999 \\
F16 F16 F16              & 67.2 & 64.6 & 64.4 & 61.7 & 2007 \\
F16 Q8 F16               & 67.7 & 65.6 & 64.7 & 59.7 & 1935 \\
Q8 ffn\_all=F16          & 72.3 & 73.7 & 70.7 & 65.5 & 1652 \\
F16 Q8 Q8                & 74.0 & 73.9 & 72.6 & 67.5 & 1652 \\
F16 F16 Q8               & 74.4 & 70.4 & 71.2 & 68.6 & 1723 \\
Q8 Q8 F16                & 75.9 & 82.6 & 78.1 & 70.5 & 1352 \\
Q8 ffn\_gate+up=F16      & 78.5 & 78.3 & 70.7 & 71.8 & 1457 \\
Q8 F16 F16               & 80.6 & 74.3 & 76.8 & 73.8 & 1424 \\
Late blocks F16          & 81.9 & 78.8 & 77.7 & 74.5 & 1396 \\
Early blocks F16         & 82.1 & 77.6 & 78.1 & 74.6 & 1396 \\
Q8 ffn\_down=F16         & 82.2 & 89.0 & 84.7 & 74.7 & 1263 \\
Q8 attn\_q+out=F16       & 84.3 & 92.3 & 87.4 & 78.7 & 1127 \\
Q8 F16 Q8                & 92.2 & 93.1 & 80.9 & 84.0 & 1141 \\
Q8 attn\_q=F16           & 92.5 & 92.5 & 84.0 & 83.0 & 1127 \\
Q8 attn\_kv+out=F16      & 92.8 & 89.7 & 86.2 & 84.7 & 1112 \\
Q8 attn\_out=F16         & 93.0 & 93.7 & 81.9 & 84.2 & 1098 \\
Q8 attn\_kv=F16          & 93.2 & 90.8 & 90.5 & 85.7 & 1084 \\
Q8 Q8 Q8                 & 95.3 & 88.3 & 90.1 & 86.2 & 1069 \\
\bottomrule
\end{tabular}
\end{table}

The results obtained justify the selection of these blocks as priority optimization targets. Thus, the search for a balance between speed gain and accuracy preservation should be focused on these segments, which served as the direct premise for applying modification metrics to them.

\textbf{Analytical latency metrics.} This work uses a simulation of uniform quantization -- an activation-quantization metric that also accounts for sensitivity, i.e., Sensitivity-Aware PTQ (SA-PTQ) -- which moves the optimization process from the realm of guesswork into the domain of precise computation and energy-based signal analysis via the SQNR coefficient. This also forgoes labor-intensive latency profiling in favor of an analytical hardware-cost metric of the layers ($L_{FFN}$, $L_{Emb}$). The integrated selectivity metric $S_n$ formed in the course of the study reveals the ``information endurance'' of each block, providing maximum speed gain under strict quality control.

\subsection{Theoretical Basis for Layer Sensitivity Estimation}

The optimization problem for language models (sLLMs) is posed by their parameter set’s information density: unlike huge models, small ones have little or no redundancy, and even minor distortions produce significant accuracy drops. The traditional Quantization-Aware Training (QAT) is prohibitively expensive for sLLMs due to requiring substantial resources for quantization~\cite{ref8,ref12}, while the canonical Post-Training Quantization (PTQ) methods~\cite{ref1,ref9,ref10,ref13,ref17,ref19}, reviewed in detail in~\cite{ref14}, apply “blind” uniform compression, destroying the connectivity structure. In this work, we propose the Sensitivity-Aware PTQ metric that evaluates a layer’s information “endurance” by quantizing its parameters and measuring the output vector’s norm deviation. By avoiding the calculation of the Hessian matrix, required for Hessian-aware mixed-precision methods~\cite{ref15} and too expensive for realistic settings, the sensitivity estimation step transforms the model into a heterogeneous structure in terms of block-wise bit-width, providing excellent speed gains while maintaining the quality within the given constraints.

The choice of PTQ as the basis for our method is motivated by the need for quick adaptation of the model to the target hardware without additional training~\cite{ref11}. Unlike traditional PTQ, which assumes uniform weight compression and is optimized to bound the parameter error, our SA-PTQ focuses on the output signal, enabling the analytical treatment of the degradation, which is crucial for the sensitivity analysis of sLLMs.

The SA-PTQ algorithm is a mathematical function that converts a smoothly varying number into a stepped approximation and vice versa. This procedure involves two operations: Quantization and Dequantization.

The step width calculation: the distance between grid points is computed based on the number of bits used to represent the numbers. For $b$ bits (e.g., $b = 4$), the grid has $2^b - 1$ intervals: For $\Delta$-bit precision (e.g., $b$), the value range of $b = 4$ is divided into $2^b - 1$ equal intervals (i.e., for 4 bits, the range would be divided into 15 equal segments):

\begin{equation}
\Delta = \frac{\max(x) - \min(x)}{2^b - 1}
\end{equation}

The step (4) serves as the unit of measurement for the entire subsequent process. Knowing it, we can derive the complete SA-PTQ mapping, which describes the number-transformation cycle:

\begin{equation}
\text{SA-PTQ}(x,b) = \text{round}\left( \frac{x - \min(x)}{\Delta} \right) \cdot \Delta + \min(x)
\end{equation}

Here, in the expression, the \texttt{round} function discards insignificant signal components, leaving only the informational skeleton corresponding to the chosen bit width. Depending on the data type inside the transformer architecture, this base logic is adapted for specific tasks.

\subsubsection{Formula variant for FFN (weight quantization)}

The FFN block (also referred to as MLP in some architectures, hence the $MLP$ subscript used in the formulas below) in the transformer architecture is traditionally viewed as the model’s ``memory’’: it is where the factual information learned by the model during training is stored. In the Gemma 3 architecture, the fully connected block is implemented as a SwiGLU variant~\cite{ref18}: instead of a standard two-layer perceptron, the output of the block is given by the element-wise product of two parallel projections, one of which is passed through a nonlinearity. While such a construction is more expressive than a standard perceptron, it becomes more challenging to analyze its sensitivity to individual matrix transformations: modifying any of the three matrices ($W_{gate}, W_{up}, W_{down}$) involved introduces non-linear distortions to the output of the FFN block.

From the point of view of quantization, the FFN block is also of particular interest: the weights of this block comprise the majority of the model’s weights, and their compression directly impacts the speed of model inference. At the same time, the hypothesis about the uniformity of sensitivity is by no means obvious: the behavior of early and late layers in the model can be significantly different.

The weights of the FFN layer are quantized using the SA-PTQ operator (4). For the per-channel scheme, the parameters $\Delta$ and $\min(W)$ are computed for each row of the matrix independently, thus minimizing the expected quantization error (5).
\begin{equation}
\widehat{W} = \text{round}\left( \frac{W_i - \min(W_i)}{\Delta_i} \right) \cdot \Delta_i + \min(W_i)
\end{equation}

where $\Delta_i = \dfrac{\max(W_i) - \min(W_i)}{2^b - 1}$ is the quantization step of row $i$. As a pessimistic baseline variant, a per-tensor scheme is considered, in which a single $\Delta$ is computed from the global maximum of the entire matrix.

To evaluate degradation at each $n$-th layer, the reference and distorted outputs of the FFN block are computed. The Gemma 3 architecture uses SiLU gating, so the output is formed as in (6).

\begin{equation}
y_{\text{clean}} = W_{\text{down}} \cdot \left( \sigma(W_{\text{gate}} h) \cdot W_{\text{up}} h \right), \qquad
y_{\text{dirty}} = \widehat{W}_{\text{down}} \cdot \left( \sigma(\widehat{W}_{\text{gate}} h) \cdot \widehat{W}_{\text{up}} h \right)
\end{equation}

where $h \in \mathbb{R}^d$ is the hidden vector of the last token, and $\sigma(x) = \dfrac{x}{1+e^{-x}}$ is the SiLU activation function. The resulting vectors are fed into the SQNR formula (7), forming a layer-wise sensitivity profile of the model with respect to FFN weight quantization. Because $y_{clean}^{(n)}$ and $y_{dirty}^{(n)}$ are computed independently per layer, the result is a full layer-wise sensitivity profile, from which the quality drop for any subset of quantized layers can be predicted.

\subsubsection{Formula variant for Embedding}

The question of quantizing language models always involves identifying the blocks that allow for aggressive compression and, therefore, are not sensitive to precision and blocks that are extremely sensitive to precision loss and, therefore, should not be compressed. The Attention block traditionally falls into the second category since the attention mechanism demonstrates fragile behavior at low bit-width and insufficient compression efficiency. As a result, this block is typically kept at a higher precision. By contrast, the Embedding matrix is a good candidate for compression since its weights are fixed and independent from the input context. Moreover, given the large vocabulary size, the embedding matrix in Gemma 3 1B constitutes a considerable portion of the model’s weights. Thus, reducing its size would lead to significant memory savings and acceleration of the model. However, this hypothesis requires empirical verification: up to which bit-width can the embedding matrix be safely compressed without affecting the quality of generated tokens? This value can be identified by quantizing the embedding matrix and measuring the tokenization quality either by direct evaluation or some proxy metric that does not require full model quantization.

The output-layer (LM Head) weights are responsible for converting the hidden state of the last token into a logit distribution over the vocabulary, from which the next token is selected. Accordingly, LM Head weights directly impact the quality of generated text: their incorrect quantization would result in a significant degradation of the logit distribution. The current work evaluates the sensitivity of the LM Head weights to low-bitwidth quantization. Notably, this analysis is performed after model training, assuming that the LM Head weights are not fine-tuned during quantization. Accordingly, the impact of reduced precision on the output-layer performance can be estimated using two proxy metrics: SQNR, which characterizes the distortion of the logit vector, and the gap metric $\Delta_{gap}$.

The output-layer weights $W$ are quantized using the SA-PTQ operator (4), which is applied to the tensor directly, as in the case of fully connected layers (5). In the per-channel variant, the parameters $\Delta$ and $\min(W)$ are computed for each row of the weight matrix, thus allowing each row to have its own dynamic range and minimizing the quantization error (7).

\begin{equation}
W_q = \text{round}\left( \frac{W_i - \min(W_i)}{\Delta_i} \right) \cdot \Delta_i + \min(W_i)
\end{equation}

where $\Delta_i = \dfrac{\max(W_i) - \min(W_i)}{2^b - 1}$ is the quantization step of row $i$. Since the weights of the Embedding matrix are static, the parameters $\Delta_i$ and $\min(W)$ are computed once and fixed for the entire inference session.

To evaluate the degradation of the output layer at each step of autoregressive generation (8), two logit vectors are computed -- the reference and the distorted one:

\begin{equation}
l_{clean} = W_{F16} \cdot h_t, \qquad l_{dirty} = W_q \cdot h_t
\end{equation}

where $W_q$ is the weight matrix after applying SA-PTQ, and $h_t \in \mathbb{R}^d$ is the hidden vector of the last token at step $t$. Feeding $l_{clean}$ and $l_{dirty}$ into the SQNR formula then gives a quantitative estimate of how much the entire logit distribution is distorted by quantizing the output-layer weights.

\subsection{SQNR as a Universal Degradation Indicator}

Within the proposed framework of activation-quantization metrics and SA-PTQ, we decided to adopt the SQNR coefficient~\cite{ref2}, which regards the compression process as the addition of noise to the information transmission channel. The main advantage of using SQNR is that the metric has a logarithmic scale and is based on energy considerations. Decibels allow to estimate distortions in both the large FFN weights and the small Embedding vectors with equal precision, while also providing a simple way to evaluate gradient of the compression ratio when reducing the bit-width (eg, from 16~to~8 bits). At the same time, SQNR is cheap to calculate for each layer, and its correlation with the final quality of generation is high, which makes it suitable for use within heterogeneous sLLMs, since these two properties are critical when selecting a metric.

The need for such a metric arises precisely from the fact that we are comparing two types of layers: FFNs and Embeddings. To do this, one needs a dimensionless coefficient that would indicate to what extent the information contained in these layers is distorted.

To do this, we use the SQNR metric (signal-to-quantization-noise ratio), which is conventionally measured in decibels. In this case, if the value of SQNR is high (eg, 40--50~dB), then this means that the layer is "not noticeable" after quantization, or, in other words, is not distorted. On the other hand, a value of SQNR below 20~dB indicates a significant loss of information during quantization. This property is very convenient for comparing different types of layers, since SQNR is a universal metric that quantifies the amount of information loss during compression, regardless of whether it is stored in vectors or tensors. The value of SQNR makes sense for any data, since it estimates the percentage of useful information in the compressed data, regardless of the initial size (whether it is millions of tokens or a hundredth of a token).

Thus, we use the SQNR (9) metric to estimate the informativeness of each layer. It calculates the ratio between the energy of information in a clean signal and the energy of information in a noisy signal. This metric helps to find the most informative layers of the neural network, which differ depending on the type of layer, within the limits of one common metric.

\begin{equation}
SQNR_{MLP}^{(n)} = 10 \cdot \log_{10}\left( \frac{\left\| y_{clean}^{(n)} \right\|_2^2}{\left\| y_{clean}^{(n)} - y_{dirty}^{(n)} \right\|_2^2} \right)
\end{equation}

The SQNR calculation for FFN measures the impact of weight quantization on the transformation of the input vector, comparing the reference value $y_{clean} = x \cdot W$ with the distorted result $y_{dirty} = x \cdot \text{SA-PTQ}(W,b)$ to assess the layer's robustness to compression, whereas the KV-cache analysis determines the direct degradation of vectors when written to memory, by comparing the original vector $y_{clean} = v$ with its quantized version $y_{dirty} = \text{SA-PTQ}(v,b)$, in order to confirm context preservation and the uniformity of the data distribution.

For a more granular view, we compute $SQNR^{(n)}$ separately for each $n$-th transformer block, which pinpoints exactly which layers are most sensitive to precision loss.

\section{Testing: Embedding Data}

\begin{figure}[H]
    \centering
    \includegraphics[width=\textwidth]{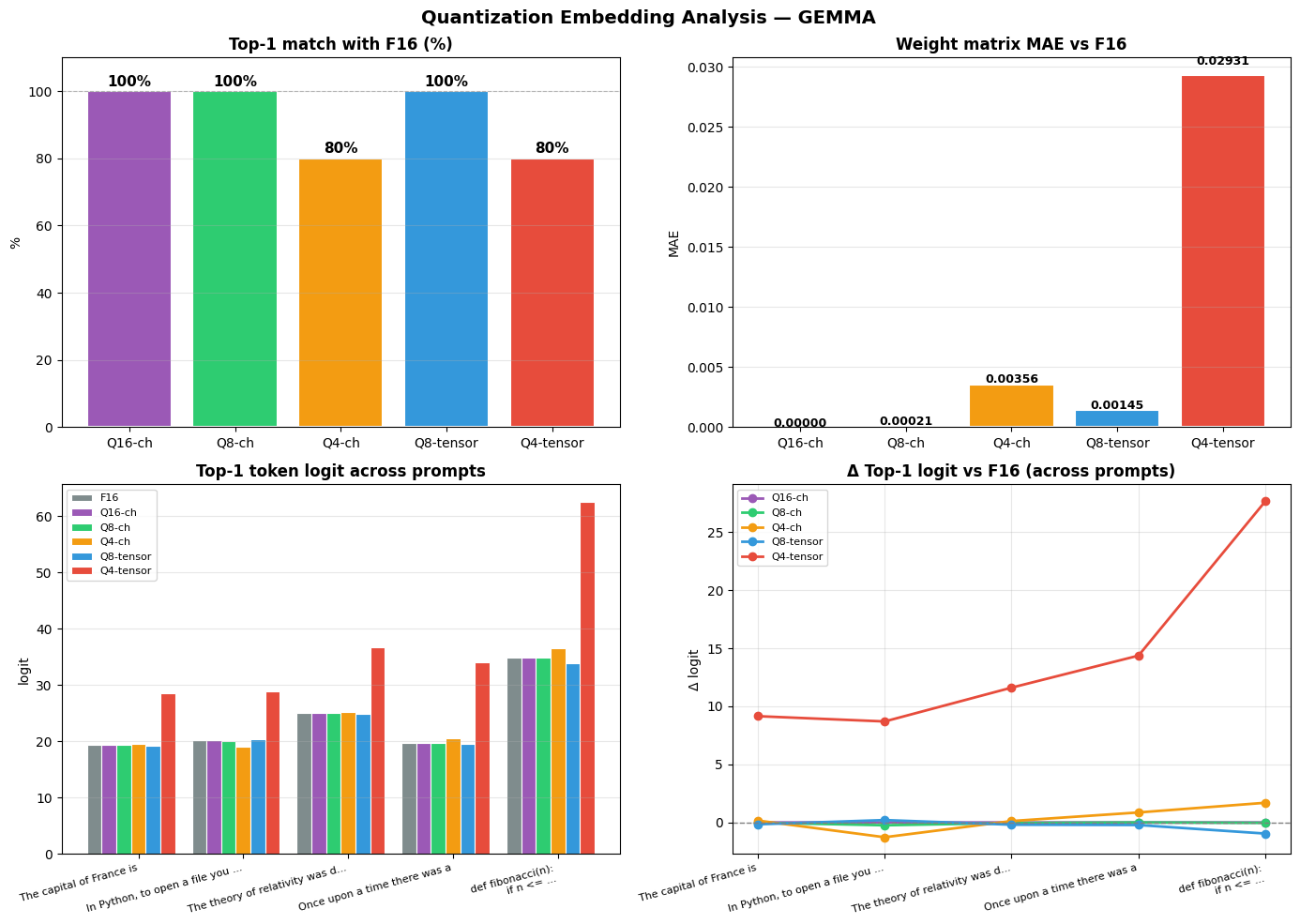}
    \caption{Token ranking for a single prompt.}
    \label{fig:top5}
\end{figure}

\begin{figure}[H]
    \centering
    \includegraphics[width=\textwidth]{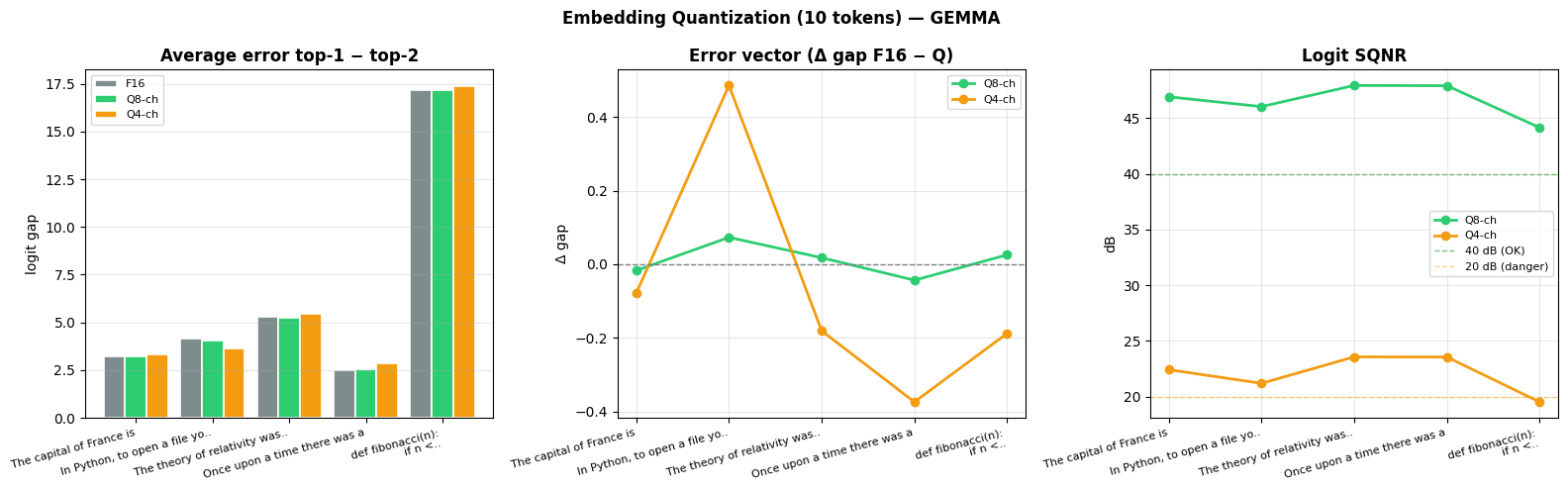}
    \caption{Difference and SQNR over 10 tokens.}
    \label{fig:diff-sqnr}
\end{figure}

Two experiments were conducted (Figs.~\ref{fig:top5},~\ref{fig:diff-sqnr}). The first (top-5 tokens for a single prompt) allows a specific moment -- a single token -- to be observed, giving a detailed picture of which tokens compete and how their ranking changes after quantization. The second (difference + SQNR over 10 tokens) allows the dynamics of generation to be observed, averaging over several steps -- a more statistically robust difference, showing how quantization affects the model's confidence during generation.

As part of an initial analysis of the degradation of the output-layer weight matrix, we perform an ablation experiment comparing model predictions at different degrees of quantization against the reference precision F16 (Fig.~\ref{fig:top5}). The ``Top-1 match with F16'' plot shows the ratio of test prompts where the winning token matched the reference model. The metric is the clearest indication of prediction degradation, as its value drops when further weight matrix reductions distort the model's final prediction.

The weight-matrix MAE is the mean absolute error between the elements of the weight matrix of the reduced precision and the reference F16 matrix. This metric characterizes the deviation of the distribution of the weights at the matrix level before they are projected to logits, enabling a direct comparison between different schemes' impact. The deviation of the value of the winning token for each prompt at each degree of reduction serves as another diagnostic metric. By plotting these values, we see ``Top-1 token logit by prompt'', the logit of the winning token for each prompt, reduced to a certain precision, versus the F16 reference. The lower value of the logit is an indication of reduced potency of this particular token in the output distribution after reduction, which could affect its rank. Similarly, delta logit of the winning token is a useful metric for inspecting individual prompts. It is calculated as the value of the logit of the winning token for the reference minus the value of the logit of the winning token for the reduced precision model, for each prompt. A positive value signifies a reduction in the winning token's potency after reduction compared to the reference, which can be a sign of prediction degradation. We analyze this value for each prompt individually, which allows us to rank prompts by their sensitivity to the reduction.

To visually assess the degradation of the output block at different quantization levels, three plots were constructed (Fig.~\ref{fig:diff-sqnr}), each reflecting a separate aspect of compression's effect on prediction quality.

The first plot shows the average gap between the top-1 and top-2 token logits, averaged over several autoregressive generation steps. This metric characterizes the model's confidence in choosing the next token: the larger the gap, the more robust the prediction is to external perturbations, including those introduced by quantization. The plot allows visual assessment of how well different bit widths preserve this confidence relative to the reference F16 model.

The second plot shows the error vector $\Delta_{gap}$ (10), the difference of the average gap between the reference and the quantized model for each test prompt. A positive value indicates a reduction in the gap between the winning token's logits and its closest competitor, which indicates degradation of the model's confidence in choosing the next token. A negative value means an increase in this difference due to the stochastic nature of quantization error and cannot be interpreted as an improvement in prediction quality. The plot makes it possible to identify prompts most vulnerable to compression and to assess the stability of degradation across different input contexts.

\begin{equation}
\Delta_{gap} = gap_{F16} - gap_{Q}
\end{equation}

The third plot shows the SQNR of the logits for each prompt. Unlike the gap metric, which focuses exclusively on the two extreme values of the distribution, SQNR evaluates the distortion of the entire logit vector. Jointly examining SQNR and $\Delta_{gap}$ makes it possible to distinguish two degradation scenarios: uniform distortion of the whole distribution versus a local shift in the region of candidate tokens that directly affects the final choice (Fig.~\ref{fig:diff-sqnr}).

For the subsequent use of SQNR in the generalized metric, in order to obtain a quantization-efficiency score, SQNR is normalized according to formula (11).

\begin{equation}
SQNR_{norm} = 1 - \frac{\log_{10}(2)}{\log_{10}(SQNR_{Q8})}
\end{equation}

The mean SQNR value for Q8-ch was 45.61 dB, giving a normalized score of 0.819. The value for Q16 (93.82 dB) serves as the reference and confirms that at full precision there is no distortion. The normalized value $Q_{emb} = \mathbf{0.819}$ reflects the information preservation of the output layer after quantization and is subsequently combined with the speed component into the unified metric.

The same normalization formula can also be applied not only to the SQNR value averaged over the entire block but also to each layer independently. Since SA-PTQ probing is performed for each layer separately, the natural output is not a single number but a vector $\{Q^{(1)}, Q^{(2)}, \ldots, Q^{(L)}\}$, where each element characterizes the information preservation of a specific layer. This opens up the possibility of layer-wise ranking, in which layers with high $Q^{(n)}$ can tolerate aggressive compression, while layers with low $Q^{(n)}$ require preserving higher precision. Thus, the proposed metric scales from a block-level estimate to a full layer-wise model profile without changing the formula itself, simply by adjusting the granularity of the input data.

\section{Testing: FFN Data}

Several tests were carried out, first, to obtain data for each FFN block (gate, up, and down) for analysis. Testing predicted the quality loss of the FFN at 8-4 bits, obtaining SQNR for each of its components and layers.

\begin{figure}[H]
    \centering
    \includegraphics[width=0.85\textwidth]{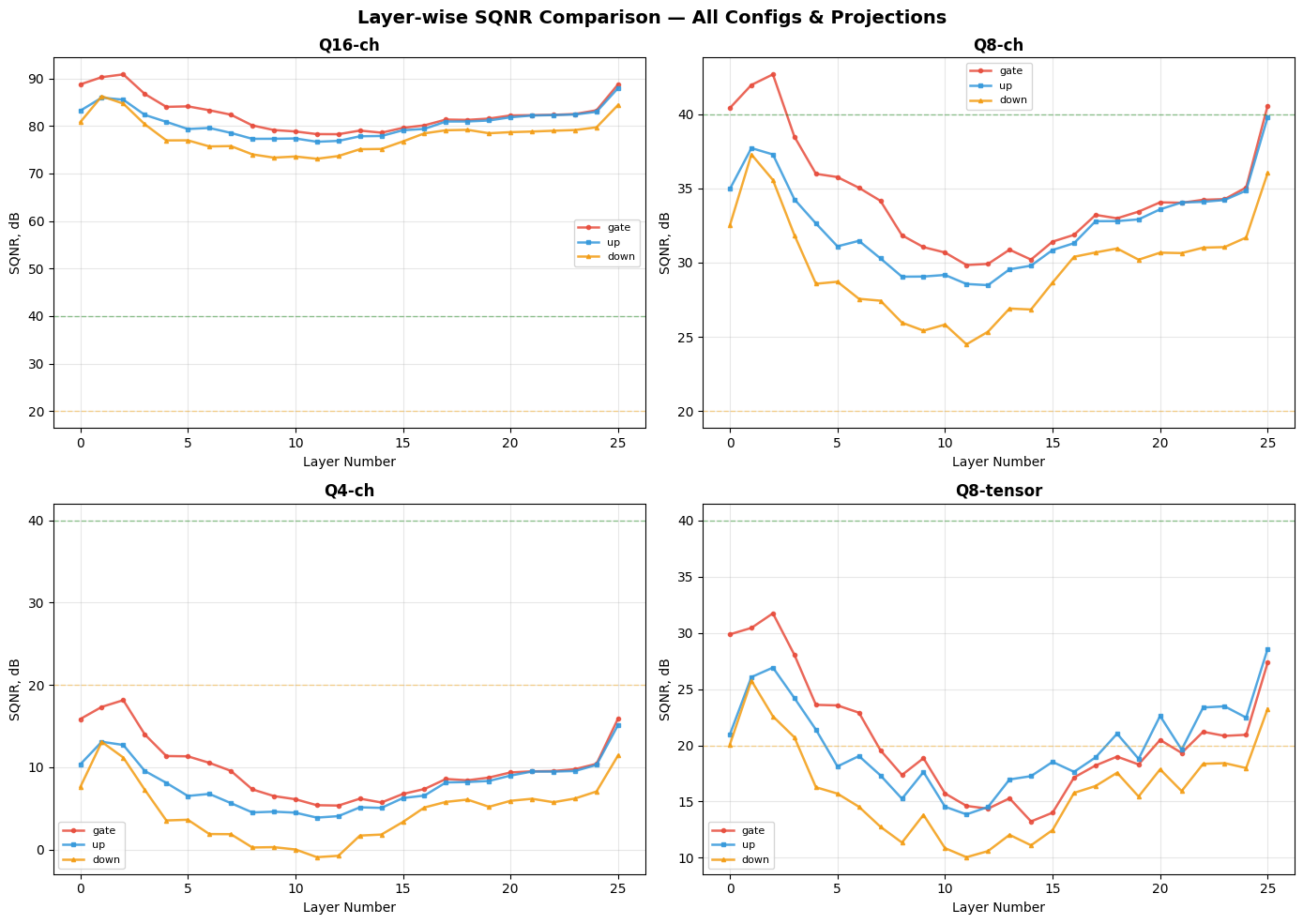}
    \caption{Predicted degradation across bit widths for each component of Gemma 3 1B-it, relative to the mean.}
    \label{fig:ffn-sqnr}
\end{figure}

\begin{figure}[H]
    \centering
    \includegraphics[width=\textwidth]{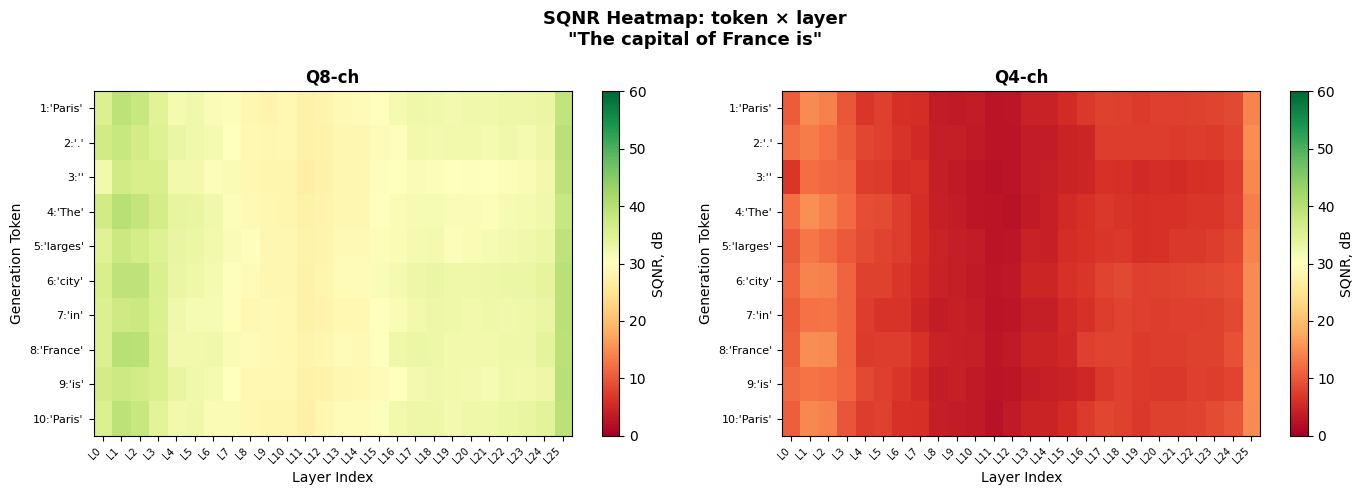}
    \caption{Per-layer/per-token prediction heatmap, geography-knowledge prompt.}
    \label{fig:heatmaps1}
\end{figure}

\begin{figure}[H]
    \centering
    \includegraphics[width=\textwidth]{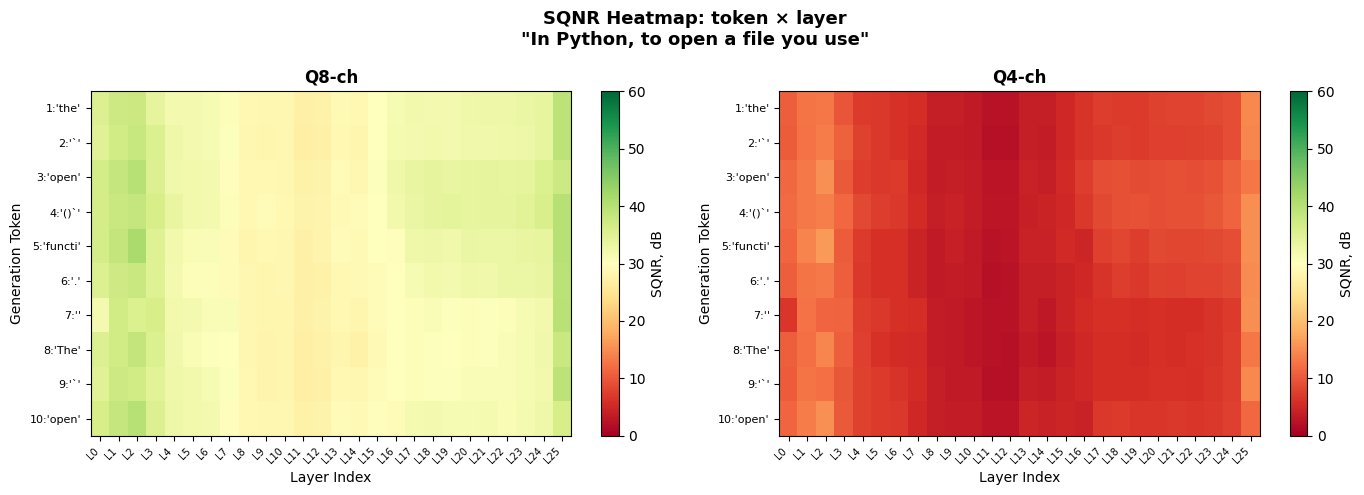}
    \caption{Per-layer/per-token prediction heatmap, function-knowledge prompt.}
    \label{fig:heatmaps2}
\end{figure}

\begin{figure}[H]
    \centering
    \includegraphics[width=\textwidth]{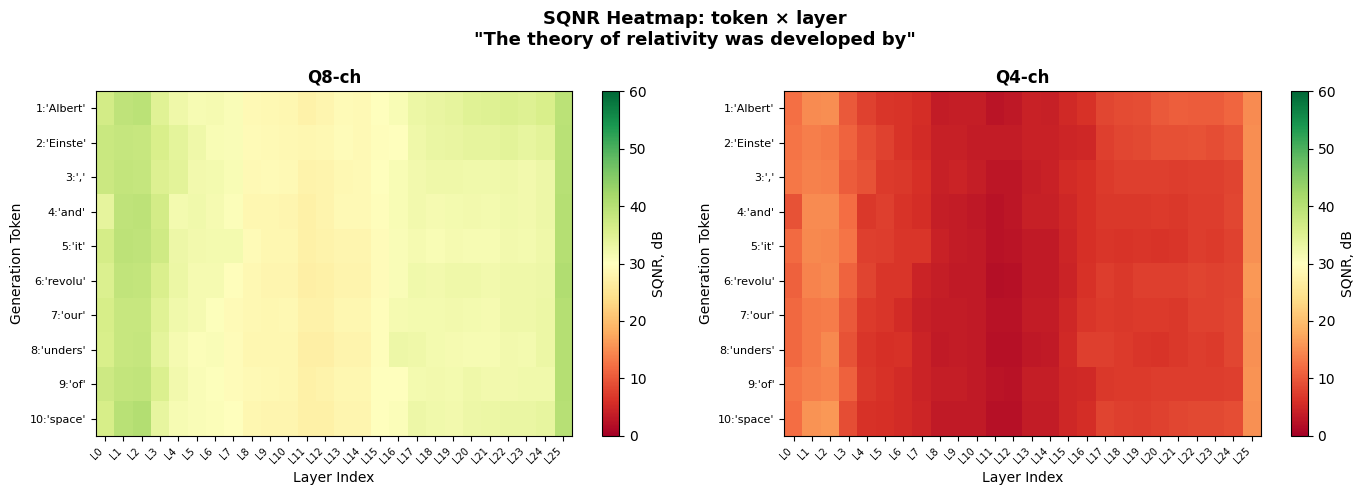}
    \caption{Per-layer/per-token prediction heatmap, historical-knowledge prompt.}
    \label{fig:heatmaps3}
\end{figure}

\begin{figure}[H]
    \centering
    \includegraphics[width=0.85\textwidth]{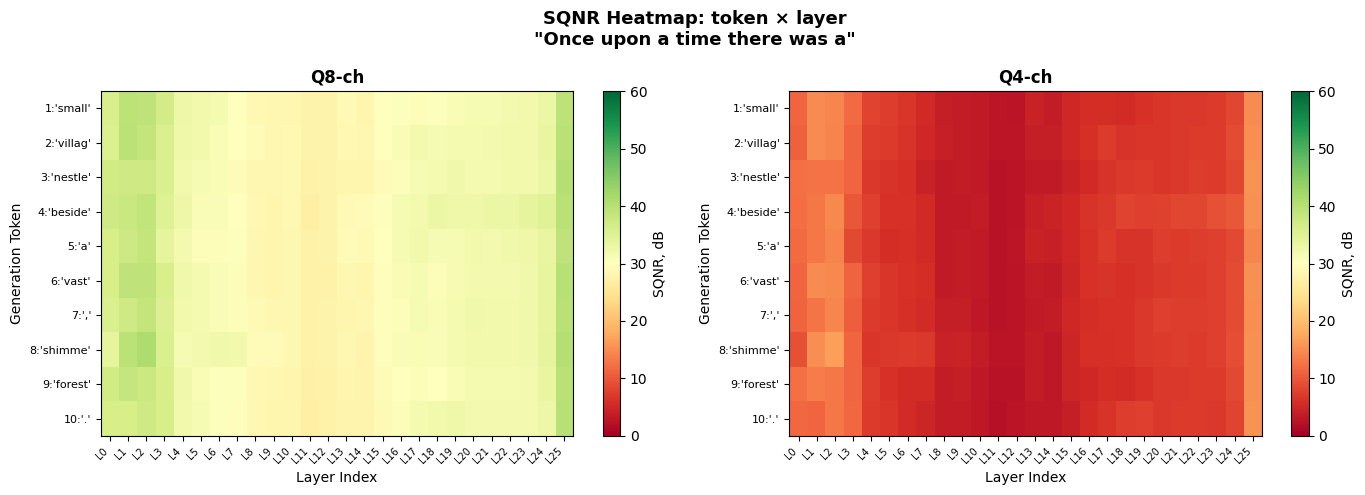}
    \caption{Per-layer/per-token prediction heatmap, composition prompt.}
    \label{fig:heatmaps4}
\end{figure}

\begin{figure}[H]
    \centering
    \includegraphics[width=0.85\textwidth]{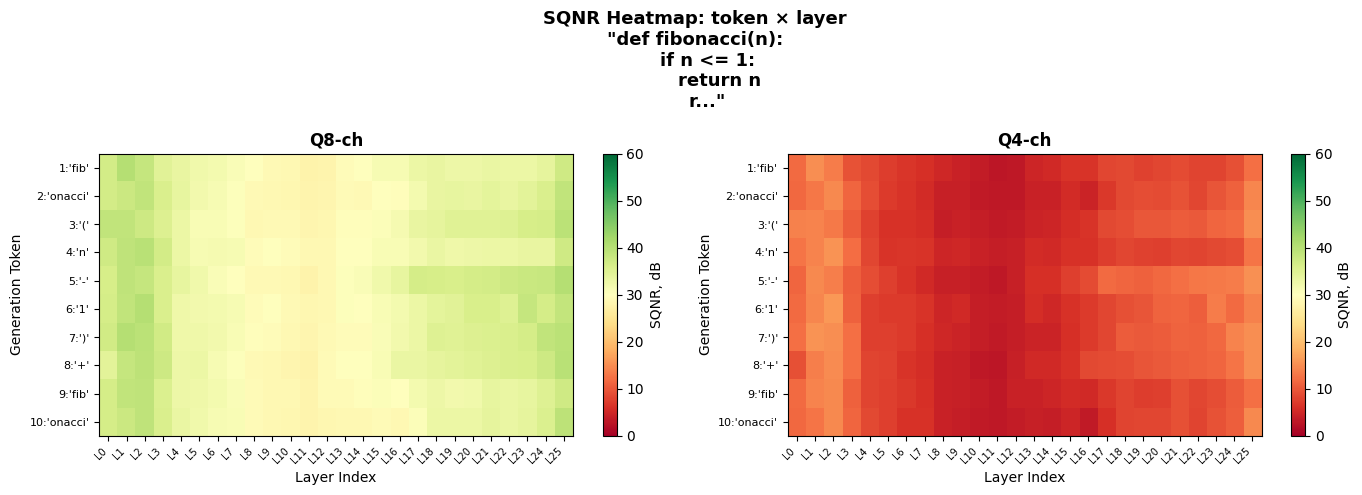}
    \caption{Per-layer/per-token prediction heatmap, code-reading prompt.}
    \label{fig:heatmaps5}
\end{figure}

Two experiments were carried out (Figs.~\ref{fig:ffn-sqnr}, \ref{fig:heatmaps1}-\ref{fig:heatmaps5}): one is a static weight analysis, yielding SQNR by projection and layer, SQNR by prompt, and weight MAE, which characterizes the general degradation picture for the given quantization scheme; the others visualize the dynamics of the autoregressive generation process by means of a token $\times$ layer heatmap, which makes it possible to see how the quantization affects each step of the generation in all layers of the model.

As part of the analysis of the degradation of the FFN block, a comparison of SQNR values was made for different levels of quantization relative to the baseline F16 (Fig.~\ref{fig:ffn-sqnr}). The ``SQNR comparison by layer - all configs and projections'' plot characterizes the dynamics of SQNR for all three projections ($W_{gate}, W_{up}, W_{down}$) for each layer of the FFN block for all four configurations. This allows one to estimate roughly the vulnerability of each projection to quantization. The drop in SQNR values in the middle layers with a tendency to rise at the output layer is characteristic of all configurations. For the Q8-ch configuration, the $W_{down}$ projection demonstrated significantly lower SQNR values compared to $W_{gate}$ and $W_{up}$, while for Q4-ch, all three projections showed SQNR values below 20 dB, which indicates the destructive effect of quantization with 4 bits per channel on the FFN block.

\begin{figure}[H]
    \centering
    \includegraphics[width=0.85\textwidth]{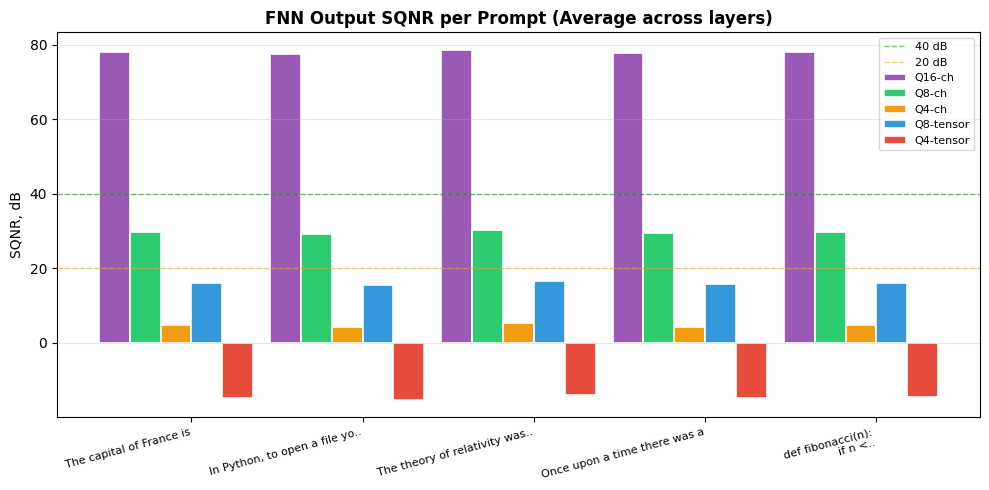}
    \caption{FFN output SQNR by prompt.}
    \label{fig:ffn-out}
\end{figure}

The plot in Fig.~\ref{fig:ffn-out} shows the mean SQNR value of the block's final output, averaged over all 26 layers of the model, for each test prompt. The results demonstrate high homogeneity of degradation: SQNR values are practically independent of the semantic content of the input context and are determined solely by the quantization scheme. Q8-ch shows a mean value of 29.0 dB -- a zone of noticeable but non-destructive degradation. Q4-ch, Q8-tensor, and Q4-tensor drop below 20 dB, with Q4-tensor showing values around 0 dB, corresponding to complete signal destruction.

\begin{figure}[H]
    \centering
    \includegraphics[width=0.85\textwidth]{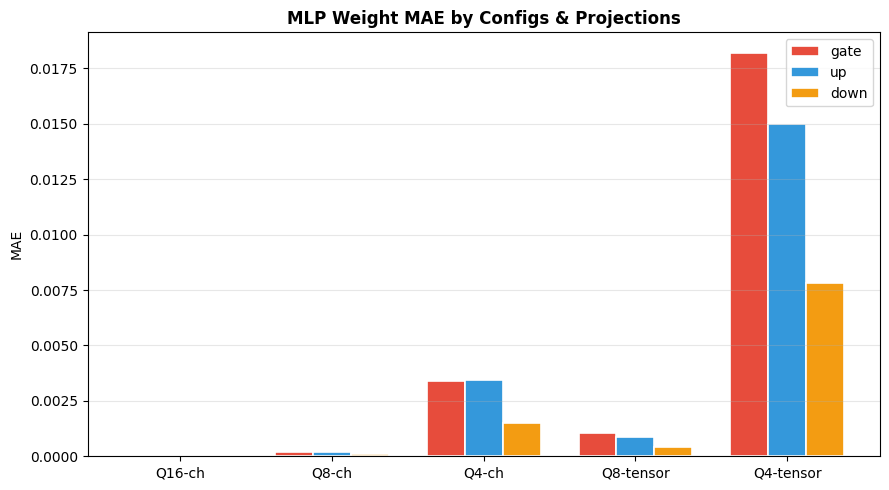}
    \caption{FFN weight MAE by configuration and projection.}
    \label{fig:ffn-mae}
\end{figure}

The plot in Fig.~\ref{fig:ffn-mae} characterizes the degree of weight distortion at the level of the matrices themselves, prior to projection into activation space. Q8-ch shows the smallest error, whereas Q4-tensor gives an order-of-magnitude larger deviation. Notably, the $W_{down}$ projection consistently shows lower MAE relative to $W_{gate}$ and $W_{up}$ across all configurations, which is explained by differences in matrix dimensionality and weight distribution.

To visually assess the dynamics of degradation during autoregressive generation, SQNR heatmaps were constructed in the token $\times$ layer space (Figs.~\ref{fig:heatmaps1}--\ref{fig:heatmaps5}). Each row corresponds to one generation step, each column to a layer number; color reflects the mean SQNR value across the three FFN projections.

Across all five prompts (Figs.~\ref{fig:heatmaps1}--\ref{fig:heatmaps5}), a consistent pattern is observed: Q8-ch forms a gradient transition from yellow-green tones in the early layers to warmer tones in the middle layers, with partial recovery toward the last layer, which correlates with the layer-wise dynamics observed in the SQNR-by-layer plot. Q4-ch, in contrast, colors nearly the entire space red, showing SQNR values below 10 dB as early as the second or third layer. The dominant degradation factor is the layer index, and this pattern is reproduced across all prompts. At the same time, context type introduces local deviations: in technical and structured prompts (``In Python, to open a file you use'', ``The theory of relativity was developed by''), individual generation tokens show anomalous SQNR values not observed in narrative contexts. This indicates that FFN block degradation is predominantly structural in nature, although the semantics of the input context can locally amplify its manifestations.

To quantitatively summarize the observed degradation picture and for subsequent use, the normalized SQNR score of the FFN block is computed using the same scheme as for the output layer (12).

\begin{equation}
SQNR_{norm} = 1 - \frac{\log_{10}(2)}{\log_{10}(SQNR_{Q8})}
\end{equation}

The mean SQNR value at Q8-ch for the $W_{down}$ projection was 29.70 dB -- noticeably lower than the corresponding value for the embedding matrix (45.61 dB), which is consistent with the visual picture from the heatmaps: the FFN block as a whole is more sensitive (since the gate, up, and down blocks are all quantized) to quantization than the output layer. The reference value Q16 (77.95 dB) confirms that at the original precision degradation is minimal. The normalized score $Q_{ffn} = \mathbf{0.796}$ reflects the higher sensitivity of the FFN to compression losses, subsequently allowing a score to be obtained based on the ratio between information loss and inference speedup.

\section{Formalizing the Latency Parameters of FFN and Embedding (Speed Component)}

To construct a general metric of quality-vs-speed dependence, the speed component must be taken into account. Since SQNR measures the ratio of signal energy to quantization-noise energy -- i.e., the degree of distortion of the output vector after weight quantization -- it is a purely informational metric that contains no information about speed. Speed is determined differently: by the number of bytes the GPU must read from memory per generation step.

To combine speed and quality in a single metric, both indicators need to be expressed on a logarithmic (decibel) scale. However, direct comparison is impossible because of the difference in scale: the speed score and SQNR lie in fundamentally different value ranges. Normalization via the logarithm -- which represents the theoretical maximum speedup for Q8 and is derived from the ratio of F16 and Q8 byte sizes -- brings both metrics to a common range. After normalization, both indicators reflect a fraction of their physical limit and can be combined into a single quantization-efficiency metric.

\subsection{Embedding}

To answer the question of how quickly the GPU can read the weight matrix from memory, only the roofline model can help -- it describes the performance of an operation through the ratio of computation to data volume. For memory-bound operations, this is sufficient, while other, more complex profiling methods or empirical measurements are redundant.

Roofline, as an analytical tool, predicts the execution time of an operation based on two physical constraints: memory bandwidth and GPU compute power. Since multiplying the embedding matrix by the hidden-state vector is memory-bound, execution time comes down entirely to memory bandwidth -- which means token time for any weight bit width can be computed analytically, from architectural parameters and hardware specs alone, without running the model at all. Applying this principle to the transformer architecture, the execution time of any layer is expressed through the size of its weights and the GPU's memory bandwidth (13).

\begin{equation}
t_{layer} = \frac{W_{bytes}}{BW}
\end{equation}

where $W_{bytes}$ is the weight size in bytes and $BW$ is the GPU's memory bandwidth (320 GB/s for the T4). The formula holds for memory-bound operations: the arithmetic intensity $AI = 1$ FLOP/byte is significantly below the ridge point of 203 FLOP/byte, hence execution time is determined exclusively by the speed of reading weights from memory. Applying this dependency to the lm\_head matrix gives its execution time for F16 and Q8 separately (14).

\begin{equation}
t_{lm}^{F16} = \frac{V \cdot H \cdot 2}{BW}, \qquad t_{lm}^{Q8} = \frac{V \cdot H \cdot 1}{BW}
\end{equation}

where $V$ is the vocabulary size and $H$ is the hidden-layer dimension. Under Q8 quantization, each matrix element occupies 1 byte instead of 2, which halves the volume of data read and proportionally speeds up the operation. Knowing the lm\_head time for both bit widths, the total time to generate one token is the sum of the time for all transformer blocks and the time for the output layer (15).

\begin{equation}
t_{tok}^{F16} = L \cdot t_{layer} + t_{lm}^{F16}, \qquad t_{tok}^{Q8} = L \cdot t_{layer} + t_{lm}^{Q8}
\end{equation}

where $L$ is the number of transformer blocks and $t_{layer}$ is the total time of all operations of one block except lm\_head. When only the embedding matrix is quantized, block time remains unchanged and only the lm\_head time changes. The ratio of times for the two configurations gives the speedup coefficient (16).

\begin{equation}
\eta_{emb} = \frac{t_{tok}^{F16}}{t_{tok}^{Q8}} = \frac{L \cdot t_{layer} + \frac{2VH}{BW}}{L \cdot t_{layer} + \frac{VH}{BW}}
\end{equation}

where $\eta$ is the speedup coefficient; the memory bandwidth $BW$ cancels out, and the speedup is expressed purely through the model's architectural parameters, independent of specific hardware characteristics. When tested on the Qwen 0.5B, Qwen 1.5B, and Gemma 3 1B models, the mean error between predicted and actual results was approximately 4\%.

To verify the proposed prediction method, a series of experiments was conducted on the Gemma 3 1B-it model using Google Colab (NVIDIA T4 GPU) as the runtime environment. Embedding-matrix quantization was performed via LLM-Viewer and llama.cpp with the Q8\_0 format, while all other model components were kept in F16. The actual speedup was measured as the ratio of the decoding speed of the quantized configuration to that of the baseline F16 model.

The results are presented in Table~\ref{tab:emb-verification}. The analytically predicted token time for Q8 was 5.37 ms versus an actual 5.51 ms, corresponding to a prediction error of 2.6\%. The predicted speedup of $\times$1.178 matched the actual $\times$1.148 within the acceptable margin of error. After normalization via $\log_{10}(2)$, the speed score was $S_{emb} = \mathbf{0.236}$, reflecting the fraction of the theoretical Q8 speedup potential that was actually realized.

\begin{table}[H]
\centering
\caption{Prediction error relative to actual speedup for Gemma 3 1B-it (Embedding).}
\label{tab:emb-verification}
\begin{tabular}{lr}
\toprule
Layers & 26 \\
\midrule
lm\_head F16 & 1.900 ms \\
lm\_head Q8 & 0.945 ms \\
Token F16 & 6.33 ms (158.0 tok/s) \\
Token Q8 (predicted) & 5.37 ms (186.1 tok/s) \\
Actual speedup & $\times$1.148, 5.51 ms (181.3 tok/s) \\
Error & 2.6\% \\
\bottomrule
\end{tabular}
\end{table}

The normalized speed score corresponds to the fraction of the theoretical Q8 speedup potential that was realized. Together with the quality component, this value forms the basis for building the unified metric.

\subsection{FFN}

For the FFN, an anchor F16 measurement is used -- a single real F16 measurement via llama-bench, which serves as the reference point for the prediction. The reasoning is that the roofline model theoretically predicts the difference between F16 and Q8 ($\Delta_{FFN}$), but not the absolute token time -- there is GPU load, the scheduler, CUDA graphs, and other factors that roofline does not account for. Since the FFN block is a heavy operation, its execution time is fully determined by the volume of weights read and by memory bandwidth (17, 18).

\begin{equation}
t_{FFN}^{F16} = \frac{W_{FFN}}{BW}, \qquad t_{FFN}^{Q8} = \frac{W_{FFN}/2}{BW}
\end{equation}

where $W_{FFN} = 3 \cdot H \cdot I \cdot 2$ bytes (gate, up, down projections). The coefficient 3 reflects the three matrices of the SwiGLU block, and the multiplier 2 is the size of one element in F16 format, in bytes. When moving to Q8, each element occupies 1 byte, so the volume of data read is halved and operation time drops proportionally. Knowing the times for both bit widths, we can compute the theoretical saving for a single block and extend it to all $L$ transformer blocks of the model (19).

\begin{equation}
\Delta_{FFN} = \left( t_{FFN}^{F16} - t_{FFN}^{Q8} \right) \cdot L
\end{equation}

The remaining components of each block -- the attention mechanism, normalizations, and residual connections -- remain in F16 and are not included in the saving computation. To move from theoretical savings to an absolute prediction, a real anchor point is needed. It is obtained as the reciprocal of the measured generation speed of the F16 model (20).

\begin{equation}
T_{real}^{F16} = \frac{1}{\text{tok/s}_{F16}}
\end{equation}

This measurement absorbs all the system overheads that roofline does not model. Given the real token time and the theoretically predicted saving, the predicted Q8 token time is obtained by subtraction (21):

\begin{equation}
T_{pred}^{Q8} = T_{real}^{F16} - \Delta_{FFN}
\end{equation}

This approach preserves the accuracy of the absolute measurement while not requiring the model to be re-run for each configuration. The ratio of the original time to the predicted time gives a dimensionless speedup coefficient (22):

\begin{equation}
\eta_{ffn} = \frac{T_{real}^{F16}}{T_{pred}^{Q8}}
\end{equation}

To evaluate the proposed prediction method, experiments were carried out on three models: Gemma 3 1B-it, LLaMA 3.2 1B, and Qwen 2.5 1.5B, using Google Colab (NVIDIA T4 GPU) as the runtime environment. For each model, the saving $\Delta_{FFN}$ was calculated analytically from the LLM-Viewer data, and the predicted speedup was compared to the actual value obtained by benchmarking with llama-bench, with only the FFN blocks being quantized to int8 and the rest of the model left in float16.

The architectural parameters and roofline-computation results for each of the three models are given in Table~\ref{tab:ffn-arch}. For Gemma 3 1B-it, the analytical saving is 1.94 ms, which would reduce the 12.53 ms per-token time to 10.59 ms. On the other hand, while LLaMA 3.2 1B has fewer blocks (16 vs. 26), its per-block saving is larger due to the higher FFN matrix width, 157.5 $\mu$s vs. 74.7 $\mu$s for Gemma 3 1B-it, for a total of 2.52 ms. Finally, with 28 blocks and the largest per-block weight volume, Qwen 2.5 1.5B has the absolute maximum saving of 3.61 ms, reducing the 17.53 ms per-token time to 13.92 ms.

\begin{table}[H]
\centering
\caption{Architectural parameters and roofline results at test time for three models.}
\label{tab:ffn-arch}
\small
\begin{tabular}{lccc}
\toprule
 & gemma-3-1b-it & llama3.2-1b & qwen2.5-1.5b \\
\midrule
Layers & 26 & 16 & 28 \\
FFN F16 (1 block) & $149.4\,\mu s \times 26 = 3.88$ ms & $314.7\,\mu s \times 16 = 5.04$ ms & $258.3\,\mu s \times 28 = 7.23$ ms \\
FFN Q8 (1 block) & $74.7\,\mu s \times 26 = 1.94$ ms & $157.5\,\mu s \times 16 = 2.52$ ms & $129.3\,\mu s \times 28 = 3.62$ ms \\
$\Delta_{FFN}$ (analytical) & 1.94 ms & 2.52 ms & 3.61 ms \\
$T_{real}^{F16}$ & 12.53 ms (79.8 tok/s) & 11.57 ms (86.4 tok/s) & 17.53 ms (57.0 tok/s) \\
$T_{pred}$ (hybrid) & 10.59 ms & 9.05 ms & 13.92 ms \\
\bottomrule
\end{tabular}
\end{table}

The comparison between predicted and actual speedup coefficients is given in Table~\ref{tab:ffn-error}. The best match was achieved for Gemma 3 1B-it, with an error of 0.7\%; for LLaMA 3.2 1B the error was 3.3\%. The largest deviation was observed for Qwen 2.5 1.5B, at 6.3\%, which may be explained by FFN architectural specifics and CUDA-scheduler behavior for this model. The mean error across the three models was 3.4\%, confirming the applicability of the hybrid approach for predicting speedup without re-running the model. After normalization, the FFN speed score was 0.261.

\begin{table}[H]
\centering
\caption{Prediction error relative to actual speedup (FFN).}
\label{tab:ffn-error}
\begin{tabular}{lccc}
\toprule
Model & Predicted & Actual & Error \\
\midrule
gemma-3-1b-it & 1.183 & 1.175 & 0.7\% \\
llama3.2-1b   & 1.278 & 1.236 & 3.3\% \\
qwen2.5-1.5b  & 1.259 & 1.344 & 6.3\% \\
\bottomrule
\end{tabular}
\end{table}

For subsequent use, the speedup coefficient is converted into a normalized scale via $\log_{10}(2)$, where a value of one corresponds to the theoretical maximum speedup at a two-fold reduction in weight volume. For Gemma 3 1B-it, the predicted speedup of $\times$1.199 gives a normalized FFN speed score of $S_{ffn} = \mathbf{0.261}$.

The presented approach admits a natural generalization to the case of layer-wise heterogeneous quantization, where each of the $L$ layers receives its own bit width $b_n$. Since all Gemma 3 layers are architecturally identical, the execution time of one layer at bit width $b_n$ is computed analytically via roofline (23).

\begin{equation}
t_{FFN,n}^{b_n} = \frac{W_{FFN} \cdot \frac{16}{b_n}}{BW}
\end{equation}

where the factor $\frac{16}{b_n}$ reflects the reduction in the volume of data read relative to F16. The total saving over all layers is defined as (24)

\begin{equation}
\Delta_{FFN} = \sum_{n=1}^{L}\left( t_{FFN,n}^{F16} - t_{FFN,n}^{b_n} \right)
\end{equation}

The predicted token time is computed by the same scheme as before. Since the architectural parameters of all 26 layers of Gemma 3 are known, the theoretical calculation for any bit-width configuration can likewise be performed analytically without additional measurements.

Considering a configuration in which the middle 10 layers -- the most sensitive according to the SQNR heatmap data -- are kept at Q8, while the first 8 and last 8 layers, being more robust, are quantized to Q4 (25, 26):

\begin{equation}
\Delta_{FFN} = 8 \cdot \left( t_{FFN}^{F16} - t_{FFN}^{Q4} \right) + 10 \cdot \left( t_{FFN}^{F16} - t_{FFN}^{Q8} \right) + 8 \cdot \left( t_{FFN}^{F16} - t_{FFN}^{Q4} \right)
\end{equation}

At a baseline token time of $t_{FFN}^{F16} = 149.4\,\mu s$, $t_{FFN}^{Q8} = 74.7\,\mu s$, $t_{FFN}^{Q4} \approx 37.4\,\mu s$:

\begin{equation}
\Delta_{FFN} = 16 \cdot 112.0 + 10 \cdot 74.7 = 1792.0 + 747.0 = 2539.0\,\mu s \approx 2.54\text{ ms}
\end{equation}

At a baseline token time of $T_{real}^{F16} = 12.53$ ms, the predicted time is $12.53 - 2.54 = 9.99$ ms, corresponding to a speedup of $\eta \approx 1.254$ versus $\eta = 1.183$ for uniform Q8. Layer-wise heterogeneity theoretically gives an additional speedup of about 6\% relative to block-level Q8, with the sensitive middle layers retaining Q8 precision while the robust outer layers are more aggressively quantized. However, it is possible that under layer-wise heterogeneous quantization the overhead of dequantizing weights when loading them into compute cores may differ for Q4 and Q8, potentially introducing an additional prediction error. Practical verification of this result requires tools with fine-grained control over the computation graph, such as ExLlamaV3 or TensorRT-LLM, which were not available for the Gemma 3 architecture within the scope of this study.

\section{The Unified Metric}

The experiments conducted for measuring quality degradation and predicting inference speed yielded four normalized values: $Q_{emb} = 0.819$, $Q_{ffn} = 0.796$, $S_{emb} = 0.236$, $S_{ffn} = 0.261$. Each of them independently characterizes one aspect of quantizing a specific block and is brought to a common scale from zero to one. The task of the final metric is to combine these four numbers into a single estimate of a quantization configuration, while preserving the ability to control the balance between speed and quality depending on the requirements of a specific task. For this, a weighted average of the speed and quality components is introduced, with a priority parameter $\alpha$ (27).

\begin{equation}
\text{score}(\alpha) = (1-\alpha)\cdot \frac{1}{N}\sum_{k} S_k + \alpha \cdot \frac{1}{N}\sum_{k} Q_k
\end{equation}

where the summation runs over all evaluated blocks $k$. In the block-level variant, $k \in \{emb, ffn\}$ and $N=2$; in the layer-wise variant, $k \in \{1, 2, \ldots, L\}$ and $N=L$. Here $S_k$ is the normalized speed score of block $k$, reflecting the fraction of the theoretical speedup potential realized at a two-fold reduction in weight volume, and $Q_k$ is the normalized SQNR score of block $k$, characterizing the information preservation of the output representations after quantization. The parameter $\alpha \in [0,1]$ sets the priority between speed and quality: at $\alpha = 0$ the metric reflects purely the speed potential, at $\alpha = 1$ purely the information preservation, and at $\alpha = 0.5$ both aspects contribute equally. All components are normalized on a unified scale, which allows quantization configurations to be directly compared both at the block level and at the level of individual layers, without reference to absolute latency values or decibels.

At $\alpha = 0.5$, for the Q8 configuration on the Gemma 3 1B-it model, the final score was $score = 0.497$, where the speed component equals 0.187 and the quality component equals 0.808.

\section{Conclusion}

This paper aims to produce a metric that quantifies the efficiency of the sLLM quantization, combining two criteria usually studied in isolation, namely, the information preserved in a given block, captured by the normalized SQNR, and the gain in the quantized model’s performance, captured by an analytical roofline latency model. Unlike prior work, where a significant computational budget was spent on either quantization-aware training or lengthy evolutionary searches over design space, the proposed metric is purely analytical, requiring no training for any of the candidate configurations.

Profiling revealed that FFN blocks and embedding contribute the most to the overall latency, with attention being surprisingly less costly, informing the choice of target blocks to optimize over. These blocks were scored according to the metrics, achieving a value of $Q_{emb} = 0.819$

and $Q_{ffn} = 0.796$ for their quality retention score, and a value of $S_{emb} = 0.236$ and $S_{ffn} = 0.261$ for their performance gain score, respectively. A sanity check on the performance gain scores across three model architectures resulted in an average error of approximately 3.4\%, demonstrating the metric’s practical applicability.

Together with the $\alpha = 0.5$, this allows us to score a particular configuration, achieving a value of 0.497 for the Q8 configuration for Gemma 3 1B-it, balancing the information preservation and performance efficiency. As demonstrated by the $\alpha$ , this metric can be adjusted to reflect the priority of either quality or performance, depending on the specific use case and system constraints. Finally, due to the metric’s independence from the granularity of input blocks, it can be applied to blocks, layers, or projections interchangeably, enabling fine-grained optimization at various levels. This property is particularly useful, as it allows one to estimate the quality drop and performance gain for a particular configuration ahead of time.

\end{document}